\documentclass[letterpaper]{article} 
\usepackage{aaai2026}  
\usepackage{times}  
\usepackage{helvet}  
\usepackage{courier}  
\usepackage[hyphens]{url}  
\usepackage{graphicx} 
\usepackage{natbib}  
\usepackage{caption} 
\usepackage{algorithm}
\usepackage{algorithmic}
\usepackage{times}
\usepackage{helvet}
\usepackage{courier}
\usepackage{xcolor}
\usepackage{amsmath}
\usepackage{amssymb}
\usepackage{graphicx} 
\usepackage{multirow}      
\usepackage{booktabs}      
\usepackage{tabularx}
\newcolumntype{Y}{>{\centering\arraybackslash}X}
\usepackage{newfloat}
\usepackage{listings}
\DeclareCaptionStyle{ruled}{labelfont=normalfont,labelsep=colon,strut=off} 
\floatstyle{ruled}
\newfloat{listing}{tb}{lst}{}
\floatname{listing}{Listing}
\title{IdealTSF: Can Non-Ideal Data Contribute to Enhancing the Performance of Time Series Forecasting Models?}
\author{
	Hua Wang\textsuperscript{\rm 1},
	Jinghao Lu\textsuperscript{\rm 1},
	Fan Zhang\textsuperscript{\rm 2}\thanks{Corresponding author}
}
\affiliations{
	\textsuperscript{\rm 1}School of Computer and Artificial Intelligence, Ludong University, Yantai, 264025, China\\
	\textsuperscript{\rm 2}School of Computer Science and Technology, Shandong Technology and Business University, Yantai, 264005, China\\
	hwa229@163.com, ljh@m.ldu.edu.cn, zhangfan@sdtbu.edu.cn
}

\usepackage{bibentry}

\begin{document}

\maketitle

\begin{abstract}
Deep learning has shown strong performance in time series forecasting tasks. However, issues such as missing values and anomalies in sequential data hinder its further development in prediction tasks. Previous research has primarily focused on extracting feature information from sequence data or addressing these suboptimal data as positive samples for knowledge transfer. A more effective approach would be to leverage these non-ideal negative samples to enhance event prediction. In response, this study highlights the advantages of non-ideal negative samples and proposes the IdealTSF framework, which integrates both ideal positive and negative samples for time series forecasting. IdealTSF consists of three progressive steps: pretraining, training, and optimization. It first pretrains the model by extracting knowledge from negative sample data, then transforms the sequence data into ideal positive samples during training. Additionally, a negative optimization mechanism with adversarial disturbances is applied. Extensive experiments demonstrate that negative sample data unlocks significant potential within the basic attention architecture for time series forecasting. Therefore, IdealTSF is particularly well-suited for applications with noisy samples or low-quality data.
\end{abstract}

\begin{links}
    \link{Code}{https://github.com/LuckyLJH/IdealTSF}
\end{links}

\section{Introduction}

Time series forecasting tasks are ubiquitous across various domains, including economics \cite{granger2014forecasting}, energy \cite{martin2010prediction, qian2019review}, transportation planning \cite{chen2001freeway, yin2021deep}, weather forecasting \cite{wu2023interpretable}, healthcare, and natural sciences. Over the past decade, deep learning methods \cite{lecun2015deep} have gained popularity in forecasting, often outperforming statistical methods such as ARIMA \cite{shumway2017arima}. However, until recently, deep learning methods for forecasting have primarily focused on selecting superior feature extraction techniques, utilizing specific training schemes, and employing foundational architectures such as CNNs \cite{wang2023micn, wu2022timesnet, hewage2020temporal}, RNNs \cite{lai2018modeling, qin2017dual}, Transformers \cite{vaswani2017attention, zhou2022fedformer, nie2022time}, and MLP variants \cite{challu2023nhits, murad2025wpmixer}, with the aim of capturing the trend characteristics of time series data as effectively as possible.

\begin{figure}[H]
	\centering
	\includegraphics[width=1\linewidth]{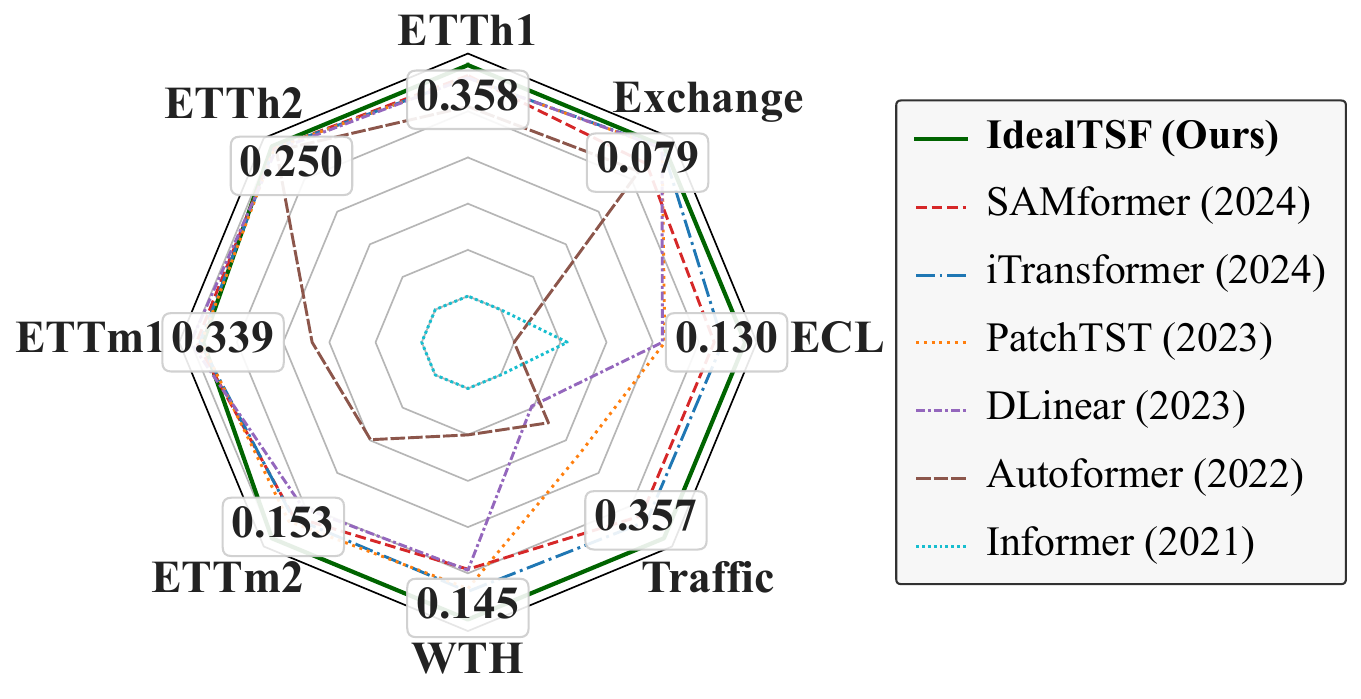}
	\caption{The performance of the model with a prediction length of 96.}
	\label{f1}
\end{figure}

Conventional time series analysis methods typically operate under the assumption that data are complete and devoid of anomalies. However, this assumption frequently fails to hold true in real-world applications \cite{schmidl2022anomaly, qu2024livescene}. Common approaches to handling missing values, such as interpolation, typically presume a linear or smooth relationship between missing entries and observed data \cite{chen2023imdiffusion}. Such methods tend to oversimplify the inherent complexity of time series data, neglecting potential nonlinear dependencies, abrupt fluctuations, and various non-stationary dynamics \cite{xu2022tgan, chen2024freegaussian}. For instance, in meteorological datasets, certain weather patterns may shift as a consequence of climate change, rendering simplistic interpolation techniques inadequate for capturing such evolving trends. In the presence of large-scale missing data, the efficacy of interpolation techniques becomes markedly limited, often failing to reconstruct the underlying data distribution accurately \cite{zeng2025FSDrive, zeng2025janusvln}. Conversely, conventional outlier detection techniques frequently depend on statistical assumptions, such as normality of the data distribution \cite{zhang2025thatsn, zhang2025probabilistic, yao2024swift}. Such assumptions may lead to the misclassification of meaningful fluctuations as anomalies, thereby compromising the predictive performance of the model. For example, in power load forecasting, seasonal patterns or exceptional events—such as national holidays—can induce substantial fluctuations in load data \cite{xiao2024confusion, wang2025c3}. Traditional anomaly detection methods may erroneously label these fluctuations as outliers, thereby overlooking critical trend information and introducing bias into forecasts \cite{xiao2025diffusion, yao2023ndc, lu2025mcnr}.

\begin{figure}[H]
	\centering
	\includegraphics[width=1\linewidth]{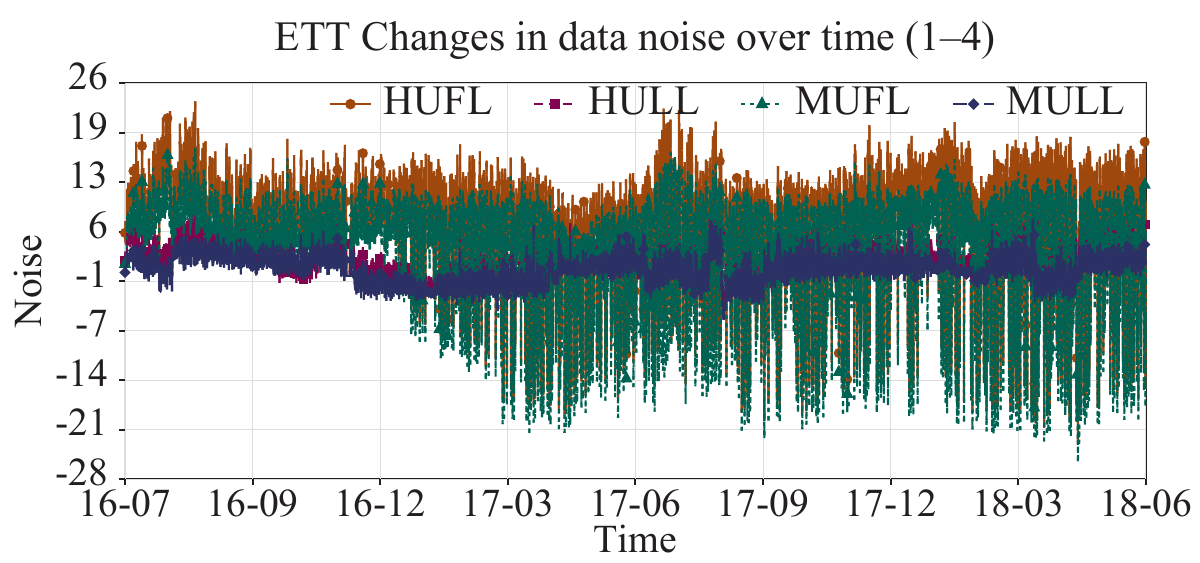}
	\caption{Irregular fluctuations in data.}
	\label{f2}
\end{figure}

In practical applications, time series data often exhibit unpredictable and non-ideal variations \cite{zhang2025enhancing, zhang2024cf}. Simply compensating for data imperfections may impair the model’s generalization ability. Therefore, enhancing the model’s robustness to imperfect conditions is critical. To this end, we propose IdealTSF, a time series forecasting framework designed to jointly strengthen the model’s resilience across the pre-training, training, and optimization phases. As illustrated in Figure \ref{f2}, the data exhibit pronounced jumps and heavy-tailed characteristics. IdealTSF explicitly addresses these non-ideal properties by introducing a negative sample pre-training module, which utilizes stable distributions to generate probabilistic data with targeted statistical features—mimicking jump processes and heavy-tailed stochastic behaviors. Additionally, it incorporates multi-scale noise injection and structured deletion to simulate anomalies and missingness commonly observed in real-world events. During training, we generate positive samples using hybrid smoothed interpolation, and integrate them with original inputs through a pre-trained attention mechanism to extract predictive features. Finally, in the optimization stage, negative perturbations are introduced to guide gradient descent toward flatter minima. To further enhance generalization, adversarial training is employed using FGSM (Fast Gradient Sign Method) or PGD (Projected Gradient Descent) attacks. This enables the model to converge more rapidly to flat optimal solutions and remain resilient to imperfect data. As demonstrated in Figure \ref{f1}, IdealTSF outperforms state-of-the-art deep models, achieving approximately a 10\% improvement in optimization metrics.

In summary, key contributions of this work are as follows:
\begin{itemize}
	\item We demonstrate that non-ideal negative samples can also provide valuable information and can be leveraged in conjunction with positive samples to collaboratively extract informative features.
	\item To fully exploit the utility of negative data, the proposed IdealTSF framework adopts a three-stage progressive design encompassing pre-training, training, and optimization phases.
	\item Extensive experiments across multiple datasets and adversarial scenarios demonstrate that the proposed model can effectively leverage negative information to resist interference from non-ideal data, achieving performance improvements exceeding 10\% over baseline methods.
	
\end{itemize}

\begin{figure*}[!t]
	\centering
	\includegraphics[width=1\linewidth]{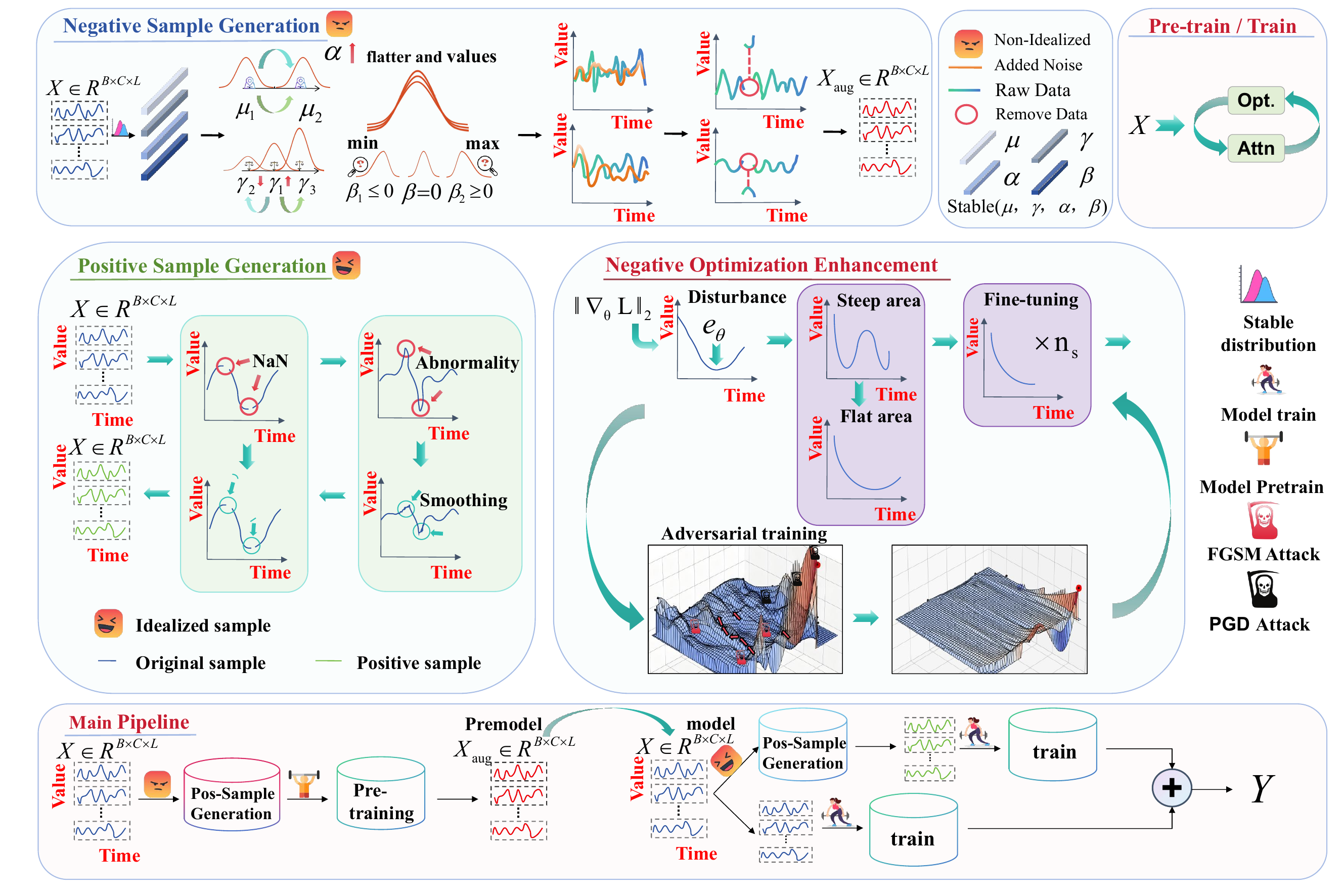}
	\caption{The architecture of IdealTSF.}
	\label{f3}
\end{figure*}

\section{Methods}
IdealTSF receives input time series data $X \in \mathbb{R}^{B \times C \times L}$, where $B$ denotes the batch size, $C$ the number of input features, and $L$ the number of time steps. The model uses historical data $X = \{X_1, X_2, ..., X_L\}$ to forecast future values $Y$. Forecasting grows harder with more variables and longer horizons, and real-world gaps/anomalies worsen it. We propose a lightweight attention model that learns from both positive and negative samples for efficient, robust prediction. Pretrain the attention module on synthetic “noisy/incomplete” data built from representative distributions, multi-scale noise, and structured deletion. Generate “clean” samples via hybrid smoothing + interpolation, then use the pretrained attention to extract features for prediction. Inject adversarial perturbations (e.g., FGSM/PGD) during training to speed convergence and improve generalization, boosting robustness to real-world imperfections. The architecture of IdealTSF is illustrated in Figure \ref{f3}.

\subsection{Negative Sample Pre-training}
To improve the model’s adaptability to imperfect data, we design a negative sample pre-training module. In each training batch, artificially perturbed negative samples are constructed from the input data $X \in \mathbb{R}^{B \times C \times L}$ to train the model and enhance its robustness.

\textbf{Stable Distributions} To simulate jump processes and heavy-tailed stochastic behaviors, we employ stable distributions as defined in Equation~\ref{Be1}, which generate probability distributions with specific statistical properties. The probability density function is determined by four parameters: the location parameter $ \mu $, the scale parameter $ \gamma $, the stability index $ \alpha $, and the skewness parameter $ \beta $. Here, $ \mu $ controls the central location of the distribution, while $ \gamma $ determines the distribution's scale or volatility. The stability index $ \alpha $ governs the tail behavior of the distribution, where $ \alpha \in (0, 2] $ and $ \alpha = 2 $ corresponds to the normal distribution. The parameter $ \beta $ controls the asymmetry of the distribution, with $ \beta = 0 $ yielding a symmetric form.

\begin{equation}
	X \sim \exp \left( i \mu t - \gamma |t|^\alpha \left( 1 - i \beta \, \text{sgn}(t) \, \tan \left( \frac{\pi \alpha}{2} \right) \right) \right)
	\label{Be1}
\end{equation}

To ensure that the generated increments are uniformly distributed across all possible directions—so that each increment has a random, rather than fixed, direction—we employ the polar coordinate method, as defined in Equation~\ref{Be2}, to generate increments from a stable distribution. This involves sampling a random angle $ \theta $ from a uniform distribution over the interval $ [0, 2\pi] $. Additionally, a uniformly distributed random variable independent of direction is sampled to maintain equal probability across all angles. The randomized angle $ \theta $ is a critical component of the polar method, as it ensures diversity in the generated negative samples by avoiding directional bias.

\begin{equation}
	\theta \sim \text{Uniform}(0, 2\pi)
	\label{Be2}
\end{equation}

Once the directional angle $\theta$ is determined, the magnitude $R$ is computed using Equation~\ref{Be3} to control the size of the increment from the stable distribution. A smaller value of the stability index $\alpha$ leads to greater volatility and jump behavior in the negative samples, effectively simulating extreme time steps. The scale parameter $\gamma$ controls the intensity of fluctuations, and the Gamma function $\Gamma(\alpha)$ is used in the computation of the distribution magnitude.

\begin{equation}
	R = \left[ \frac{\gamma}{2} \left( \frac{|\Gamma(\alpha)|}{1 - \alpha} \right) \right]^{\frac{1}{\alpha}}
	\label{Be3}
\end{equation}

After obtaining the increment magnitude $R$, it is combined with the random angle $\theta$ using Equation~\ref{Be4} to generate the final increment $\Delta x_i$. This step ensures that the increment is random in both magnitude and direction, thereby capturing the jump characteristics inherent in the data. The resulting negative sample sequence is constructed using Equation~\ref{Be5}, yielding $X'(T) = { x'_1, x'_2, \dots, x'_n }$.

\begin{equation}
	\Delta x_i = R \cdot \cos(\theta)
	\label{Be4}
\end{equation}

\begin{equation}
	x_i' = x_i + \Delta x_i
	\label{Be5}
\end{equation}

\textbf{Multi-scale Noise} In real-world applications, time series data may experience various abrupt perturbations. To enhance the model's adaptability to such scenarios, we introduce multi-scale noise during the negative sample pre-training phase. Specifically, disturbances of varying frequencies are simulated by adding noise at different scales $ { w_i } $. Noise $ n(t) $ is added to the perturbed time series $ X'(T) $ using Equation~\ref{e1}, thereby improving model robustness. For each scale $ w_i $, the corresponding noise $ n(t) $ is generated via Equation~\ref{e2} with a distinct noise intensity $ \sigma_i $, where $ \mathcal{N}(0, \sigma_i^2) $ denotes a standard normal distribution representing the noise at each time step. A sliding window operation is applied to the noise across different scales to simulate the blurring effects of multi-scale perturbations. Lower-frequency (longer time-scale) noise is assigned higher intensity, while higher-frequency (shorter time-scale) noise is assigned lower intensity.

\begin{equation}
	x_{\text{noise}}(t) = X'(T) + n(t),
	\label{e1}
\end{equation}

\begin{equation}
	n(t) = \frac{1}{w_i} \sum_{\tau = t - w_i + 1}^{t} \mathcal{N}(0, \sigma_i^2)
	\label{e2}
\end{equation}

\textbf{Structured Deletion} To further improve robustness to missing or irregular data, the model adopts a structured deletion strategy by randomly removing continuous segments from negative samples to simulate real-world data loss. For the input $x_{\text{noise}}(t)$, data is deleted over a randomly selected time interval using Equation~\ref{e3}, where $L$ denotes the length of the deletion segment, and $L \in [L_{\min}, L_{\max}]$. Within the deleted interval, time series values are set to zero. Here, $t_s$ represents the start time of deletion, and $L_d \sim \mathcal{U}(L_{\min}, L_{\max})$ denotes the deletion length sampled from a uniform distribution. The final augmented sequence $x_{\text{aug}}$ is obtained using Equation~\ref{e4}.

\begin{equation}
	x_{\text{aug}}(t) =
	\begin{cases}
		0, & t \in [t_s, t_s + L_d] \\
		x_{\text{noise}}(t), & \text{otherwise}
	\end{cases}
	\label{e3}
\end{equation}

\begin{equation}
	x_{\text{aug}} = [x_{\text{aug}}(1), ..., x_{\text{aug}}(L)]
	\label{e4}
\end{equation}

\subsubsection{Negative Sample Training} We pre-train the attention module by minimizing the mean squared error (MSE) between the model output and the ground truth labels, as defined in Equation~\ref{bu}. Additionally, we incorporate the EcoSystem Optimizer (ECOS) to enhance the stability of parameter updates via negative enhancement strategies. Here, $\theta$ denotes the trainable parameters of the model, $\mathcal{L}_{\text{MSE}}$ is the mean squared error loss function, and $f_{\theta}(X)$ represents the model prediction based on input $X$ and parameters $\theta$.

\begin{equation}
	\theta^* =
	\begin{cases}
		\arg\min_\theta \, \mathbb{E}_{(X, Y)}\left[\mathcal{L}_{\text{MSE}}(f_{\theta}(X), Y)\right] \\
		\text{s.t.} \quad \theta $←$ \text{ECOS}(\nabla_\theta \mathcal{L}_{\text{MSE}})
	\end{cases}
	\label{bu}
\end{equation}

\subsection{EcoSystem Optimizer (ECOS)}

The stability of deep learning models depends heavily on a robust optimization environment. Inspired by the resilience of ecosystems—where recovery is possible even after natural disasters—we propose the EcoSystem Optimizer (ECOS). This method simulates external disturbances using adversarial attacks such as FGSM or PGD, while internal perturbations are introduced by injecting noise into the gradient descent process. Together, these mechanisms enhance the robustness and generalization ability of time series forecasting models.

\textbf{Adversarial Sample Generation} Before optimization, adversarial samples $\mathbf{x}_{\text{adv}}$ are generated using adversarial training techniques such as FGSM or PGD. For each input $\mathbf{x}$ and its corresponding label $\mathbf{y}$, the perturbation is computed according to Equation~\ref{en1}, where $\alpha$ is the perturbation step size, and $\nabla{\mathbf{x}} L(\mathbf{x}, \mathbf{y})$ is the gradient of the loss function with respect to the input, guiding the direction of the adversarial modification. Furthermore, Equation~\ref{en2} ensures that the generated adversarial sample remains within a bounded perturbation region $\epsilon$ around the original input $\mathbf{x}$.

\begin{equation}
	\mathbf{x}_{adv} = \mathbf{x} + \alpha \cdot \text{sign}(\nabla_{\mathbf{x}} L(\mathbf{x}, \mathbf{y}))
	\label{en1}
\end{equation}

\begin{equation}
	\mathbf{x}_{adv}^{k+1} = \text{clip}_{\mathbf{x}, \epsilon} \left( \mathbf{x}_{adv}^{k} + \alpha \cdot \text{sign} \left( \nabla_{\mathbf{x}} L (\mathbf{x}_{adv}^{k}, y) \right) \right)
	\label{en2}
\end{equation}

\textbf{Phase I: Internal Perturbation Resistance and Flat Region Exploration} After completing adversarial preparation, the goal of the first phase is to enhance the model’s robustness against internal perturbations by “ascending” the loss landscape—adjusting parameters toward a local maximum. First, the gradient of the loss with respect to the model parameters is computed using Equation~\ref{en3}, yielding $\nabla_\theta L$. Then, perturbation $e_\theta$ is generated via Equation~\ref{en4}, where $\rho$ controls the magnitude of the perturbation. Gradient normalization is applied to ensure stable updates. Finally, the perturbation is applied to the parameters using Equation~\ref{en5}, improving the model’s adaptability to previously unseen samples.

\begin{equation}
	\left\| \nabla_\theta L \right\|_2 = \sqrt{\sum_i (\nabla_{\theta_i} L)^2}
	\label{en3}
\end{equation}

\begin{equation}
	e_\theta = \frac{\rho}{\left\| \nabla_\theta L \right\|_2} \cdot \nabla_\theta L
	\label{en4}
\end{equation}

\begin{equation}
	\theta_{\text{new}} = \theta + e_{\theta}
	\label{en5}
\end{equation}

\textbf{Phase II: Multi-step Fine-tuning} After the parameter updates in the first phase, the model proceeds to a multi-step fine-tuning stage. A base optimizer (e.g., Adam) is used to perform $n_s$ small-step updates on each parameter with a learning rate $\eta$. In each step, forward propagation is conducted, gradients $\nabla L$ are computed, and backpropagation is applied for parameter adjustment. This iterative process avoids large updates, progressively guides the model toward a better solution, and enhances both training stability and convergence performance.

\begin{equation}
	\theta_i^s = \theta_i^{s-1} - \eta \cdot \nabla L(\theta_i^{s-1})
	\label{en6}
\end{equation}

\textbf{Phase III: Parameter Restoration and Base Optimization} In the second phase, the parameters $\theta$ are restored to their pre-perturbation state using Equations~\ref{en7}–\ref{en8}, followed by a standard optimization update. Although this step is referred to as a “restoration,” the actual objective is not to return to the original point, but rather to guide the parameters toward a flatter region in the loss landscape, thereby reducing the risk of falling into sharp local minima.

\begin{equation}
	\theta_{\text{recovered}} = \theta_{\text{new}} - e_\theta
	\label{en7}
\end{equation}

\begin{equation}
	\theta_{\text{final}} = \theta_{\text{recovered}} - \eta \nabla_\theta L
	\label{en8}
\end{equation}

\textbf{Adversarial Training} In the final step of optimization, ECOS performs forward propagation using the adversarial samples $\mathbf{x}_{\text{adv}}$, and computes the mean squared error loss $L(\mathbf{x}_{\text{adv}}, y)$ as defined in Equation~\ref{en9}. Subsequently, gradients are computed via backpropagation using Equation~\ref{en10}, and the parameters are updated through the base optimizer to ensure that the model is effectively optimized even on adversarial examples.

\begin{equation}
	L_{\text{adv}} = L({x}_{adv}, y)
	\label{en9}
\end{equation}

\begin{equation}
	\nabla_\theta L_{\text{adv}} = \frac{\partial L_{\text{adv}}}{\partial \theta}
	\label{en10}
\end{equation}

\begin{equation}
	\theta = \theta - \eta \nabla_\theta L_{\text{adv}}
	\label{en11}
\end{equation}	

\subsection{Pre-training → Positive Sample Training} In the preceding stage, negative sample pre-training enabled the model to adapt to imperfect data conditions. In the formal training phase, we addressed potential data anomalies to guide the model in learning the normal patterns of the time series.

\textbf{Missing Data Detection} First, the missing data points $T_{\text{missing}}$, representing unavailable time steps, are identified. This set is defined by Equation~\ref{e11}, and the missing values are marked using $NaN$.

\begin{equation}
	T_{\text{missing}} = \{t \mid x(t) = \text{NaN}\}
	\label{e11}
\end{equation}

\textbf{Anomaly Detection} A hybrid anomaly detection approach is employed to identify outliers. First, the Z-score is calculated using Equation~\ref{e12} to measure the deviation of data from the mean $\mu$; if the absolute value exceeds a predefined threshold $\alpha$, the point is flagged as a preliminary anomaly. Additionally, the interquartile range is computed using Equation~\ref{e13} to detect further anomalies, where $Q1$ and $Q3$ represent the lower and upper quartiles, respectively. 
\begin{equation}
	Z(t) = \frac{x(t) - \mu}{\sigma}
	\label{e12}
\end{equation}

\begin{equation}
	\text{IQR} = Q3 - Q1
	\label{e13}
\end{equation}

According to Equation~\ref{e14}, anomalous data points are identified using a combined approach based on the Z-score and the Interquartile Range (IQR) methods. The threshold $\wp$ is typically set to 2 or 3. While the Z-score is suitable for normally distributed data, the IQR value $I$ is more appropriate for non-normal distributions. This dual-criteria strategy enhances the model’s ability to accurately detect outliers.

\begin{equation}
	x(t) =
	\begin{cases}
		\text{Abnormal}, & 
		\begin{aligned}
			\text{if } \; & |Z(t)| > \wp \quad \text{or}\\
			&  x(t) < Q1 - 1.5 \times \text{I} \quad \text{or} \\
			&  x(t) > Q3 + 1.5 \times \text{I}
		\end{aligned} \\
		\text{Correct}, & \text{otherwise}
	\end{cases}
	\label{e14}
\end{equation}

\textbf{Positive Sample Generation} To better leverage the robustness of the pre-trained attention mechanism, linear spline interpolation is employed to fill long-duration missing segments. For each missing time step $t \in T_{\text{missing}}$, interpolation is performed using Equation~\ref{e15}, which estimates the missing value based on the nearest known time steps $t_1$ and $t_2$. In addition, the interpolated results are smoothed using Equation~\ref{e16} to eliminate short-term fluctuations, where $W$ denotes the size of the sliding window. By applying the above hybrid interpolation methods to fill in missing values and correct anomalies, the complete time series is ultimately reconstructed using Equation~\ref{e17}. The resulting $\hat{x}$ serves as the positive sample for subsequent model training.

\begin{equation}
	x(t) = x(t_1) + \frac{t - t_1}{t_2 - t_1} \cdot (x(t_2) - x(t_1)), \quad t \in [t_1, t_2]
	\label{e15}
\end{equation}

\begin{equation}
	\hat{x}(t) = \frac{1}{W} \sum_{t' = t - W + 1}^{t} x(t')
	\label{e16}
\end{equation}

\begin{equation}
	x_{\text{aug}} = [\hat{x}(1), \hat{x}(2), ..., \hat{x}(L)]
	\label{e17}
\end{equation}

\textbf{Dual-Channel Feature Capture} As shown in Equation~\ref{e20}, the original time series $x_{\text{orig}}$ and the generated positive sample $x_{\text{aug}}$ are combined as a dual-channel input and fed into the pre-trained attention mechanism. This design aims to demonstrate that, under complete data conditions, even a basic attention mechanism can achieve strong performance.

\begin{equation}
	z_{\text{orig}} = [x_{\text{orig}}(1), x_{\text{orig}}(2), ..., x_{\text{orig}}(L)] \in \mathbb{R}^{B \times C \times L}
	\label{e18}
\end{equation}

\begin{equation}
	z_{\text{aug}} = [x_{\text{aug}}(1), x_{\text{aug}}(2), ..., x_{\text{aug}}(L)] \in \mathbb{R}^{B \times C \times L}
	\label{e19}
\end{equation}

\begin{equation}
	Attention_{train}
	\begin{cases}
		z_{\text{orig}} \\
		z_{\text{aug}}
	\end{cases}
	\label{e20}
\end{equation}

\section{Experiments} In this section, we present the experiments of IdealTSF on multiple mainstream datasets and the effectiveness tests of relevant modules. To prove the robustness of IdealTSF against non - ideal real - world factors, we conducted experiments on all baseline models in short - term prediction by randomly adding noise ranging from 5\% to 35\% again. Meanwhile, to ensure fairness, we presented the complete experimental data without noise addition in the appendix.

\textbf{Benchmarks}  We conducted extensive experiments to evaluate the performance and efficiency of IdealTSF, covering both long-term and short-term forecasting tasks. For long-term forecasting, we conduct experiments on eight benchmark datasets: the ETT series—including four subsets (ETTh1, ETTh2, ETTm1, and ETTm2)—as well as Weather, Electricity, and Traffic datasets. For short-term forecasting, we evaluate on the PEMS dataset.

\begin{table*}[t]
	\centering
	\setlength{\tabcolsep}{2pt}
	\begin{tabularx}{\textwidth}{l c c c | *{7}{>{\centering\arraybackslash}X X}}
		\toprule
		\textbf{} & \multicolumn{1}{c}{} & \multicolumn{2}{c|}{Ours} & \multicolumn{6}{c|}{2025}   & \multicolumn{8}{c}{2023-2024}  \\
		\midrule
		\textbf{Models} & \multicolumn{1}{c}{H} & \multicolumn{2}{c|}{IdealTSF} & \multicolumn{2}{c}{Twins*} & \multicolumn{2}{c}{Time**}& \multicolumn{2}{c}{TimeKAN}& \multicolumn{2}{c}{SAM*} & \multicolumn{2}{c}{iTrans*} & \multicolumn{2}{c}{PatchTST} & \multicolumn{2}{c}{DLinear}  \\
		\cmidrule(lr){3-4} \cmidrule(lr){5-6} \cmidrule(lr){7-8} \cmidrule(lr){9-10} \cmidrule(lr){11-12} \cmidrule(lr){13-14} \cmidrule(lr){15-16} \cmidrule(lr){17-18}
		& & MSE & MAE & MSE & MAE & MSE & MAE & MSE & MAE & MSE & MAE & MSE & MAE & MSE & MAE & MSE & MAE \\
		\midrule
		\multirow{1}{*}{ETTh1}
		& Avg & \textbf{0.402} & \textbf{0.419} & {0.446} &{0.440} & 0.419 & 0.432 & 0.417 & 0.427 &0.432 &0.424 &0.454 &0.448 & 0.438 & 0.449 & 0.441 & 0.439 \\
		\midrule
		\multirow{1}{*}{ETTh2}
		& Avg & \textbf{0.338} & {0.393} & {0.373} &{0.400} & 0.339 & \textbf{0.380} & 0.383 & 0.404 &0.344 &0.392 &0.383 &0.407 & 0.384 & 0.414 & 0.548 & 0.521 \\
		\midrule
		\multirow{1}{*}{ETTm1}
		& Avg & {0.409} & {0.431} & {0.393} &{0.404} & \textbf{0.369} & \textbf{0.378} & 0.377 & 0.395 &0.373 &0.388 &0.407 &0.410 & 0.391 & 0.412 & 0.400 & 0.412 \\
		\midrule
		\multirow{1}{*}{ETTm2}
		& Avg & \textbf{0.248} & \textbf{0.302} & {0.277} &{0.323} & 0.269 & 0.320 & 0.277 & 0.323 &0.269 &0.327 &0.288 &0.332 & 0.280 & 0.316 & 0.350 & 0.392 \\
		\midrule
		\multirow{1}{*}{WTH}
		& Avg & \textbf{0.201} & \textbf{0.249} & {0.246} &{0.271} & 0.226 & 0.262 & 0.243 & 0.272 &0.261 &0.293 &0.258 &0.278 & 0.261 & 0.280 & 0.274 & 0.349 \\
		\midrule
		\multirow{1}{*}{Traffic}
		& Avg & \textbf{0.371} & \textbf{0.253} & {0.407} &{0.274} & 0.416 & 0.264 & 0.422 & 0.269 &0.425 &0.297 &0.428 &0.282 & 0.555 & 0.395 & 0.632 & 0.397 \\
		\midrule
		\multirow{1}{*}{ECL}
		& Avg & \textbf{0.156} & \textbf{0.252} & {0.167} &{0.261} & 0.165 & 0.253 & 0.182 & 0.274 &0.181 &0.275 &0.176 &0.270 & 0.209 & 0.306 & 0.211 & 0.303 \\
		\midrule
		\multirow{1}{*}{PEMS03}
		& Avg & \textbf{0.106} & \textbf{0.208} & {0.109} &{0.219} & 0.116 & 0.226 & --- & --- &0.180 &0.304 &0.169 &0.272 & 0.376 & 0.329 & 0.159 & 0.262 \\
		\midrule
		\multirow{1}{*}{PEMS04}
		& Avg & \textbf{0.102} & \textbf{0.202} & {0.111} &{0.219} & 0.121 & 0.232 & --- & --- &0.195 &0.307 &0.209 &0.317 & 0.353 & 0.420 & 0.130 & 0.241 \\
		\midrule
		\multirow{1}{*}{PEMS07}
		& Avg & 0.118 & {0.220} & \textbf{0.094} &\textbf{0.196} & 0.100 & 0.204 & --- & --- &0.211 &0.303 &0.235 &0.315 & 0.380 & 0.440 & 0.125 & 0.226 \\
		\midrule
		\multirow{1}{*}{PEMS08}
		& Avg & {0.182} & {0.250} & \textbf{0.133} &\textbf{0.222} & 0.151 & 0.234 & --- & --- &0.280 &0.321 &0.268 &0.306 & 0.440 & 0.363 & 0.192 & 0.271 \\
		
		\bottomrule
	\end{tabularx}
	\footnotesize * indicates Former; ** indicates Mixer++.
	\caption{Average performance on both long-term and short-term time series forecasting tasks. Bold indicates the best result. The above results are the average values of 5 different random seeds.}
	\label{t2}
\end{table*}

\subsection{Long/Short-Term Time Series Forecasting}
Table \ref{t2} presents a comparison between IdealTSF and other baseline models across eleven benchmark datasets. The results show that IdealTSF consistently ranks among the top two on all datasets, often achieving or closely approaching state-of-the-art performance. These datasets span various series with different sampling frequencies, numbers of variables, and real-world application scenarios. Notably, IdealTSF significantly outperforms models like Twinformer \cite{zhoutwinsformer}, TimeMixer++ \cite{wang2024timemixer++}, TimeKAN \cite{huang2025timekan}, SAMformer \cite{ilbert2024samformer}, and PatchTST \cite{nie2022time}, and, relative to TimeKAN, reduces MSE on ECL and ETTh1 by approximately 17\% and 3.5\%, respectively. Importantly, IdealTSF also demonstrates robust performance on datasets with inherently low predictability, such as Traffic and ECL, further validating its generalization ability. In addition to long-term forecasting, IdealTSF achieves strong results in short-term forecasting tasks as well. On the PEMS benchmark—which consists of multiple time series recorded across a city-wide traffic network—many advanced models (e.g., PatchTST (2023) and DLinear (2023)) exhibit performance degradation due to complex spatiotemporal dependencies among variables. In contrast, IdealTSF maintains competitive performance on this challenging task, highlighting its effectiveness in modeling complex multivariate time series.

\subsection{Ablation Study}
To evaluate the effectiveness of each component in IdealTSF, we conducted ablation studies by systematically removing individual modules (denoted as w/o). Table~\ref{t5} presents detailed results and corresponding analysis. Specifically, "w/o Neg" refers to the removal of the negative sample pre-training module, "w/o Pos" denotes the removal of the positive sample generation process, and "w/o ECOS" indicates the exclusion of the Ecosystem Optimizer. "w/o Pos+ECOS" removes both the positive sample training and ECOS, while "w/o Neg+ECOS" removes both negative sample pre-training and ECOS. From the MSE and MAE metrics across multiple datasets (ETTh1, ETTh2, ETTm1, and ETTm2), it is evident that IdealTSF consistently outperforms the ablated versions, demonstrating superior forecasting accuracy and stability. These results confirm the contribution of each module to the overall performance and robustness of the model. 

\begin{table*}[t]
	\centering
	\setlength{\tabcolsep}{2pt}
	\begin{tabularx}{\linewidth}{c|c|*{12}{>{\centering\arraybackslash}X}}
		\toprule
		\multicolumn{2}{c|}{\textbf{Models}}  & \multicolumn{2}{c|}{\textbf{IdealTSF}} & \multicolumn{2}{c|}{\textbf{w/o Neg}} & \multicolumn{2}{c|}{\textbf{w/o Pos}} & \multicolumn{2}{c|}{\textbf{w/o ECOS}}   & \multicolumn{2}{c|}{\textbf{w/o Pos+ECOS}} & \multicolumn{2}{c}{\textbf{w/o Neg+ECOS}} \\
		\multicolumn{2}{c|}{\textbf{Metric}} & \textbf{MSE} & \textbf{MAE} & \textbf{MSE} & \textbf{MAE} & \textbf{MSE} & \textbf{MAE} & \textbf{MSE} & \textbf{MAE} & \textbf{MSE} & \textbf{MAE} & \textbf{MSE} & \textbf{MAE} \\
		\midrule
		\multirow{1}{*}{\textbf{ETTh1}} 
		& Avg  & \textbf{0.402} & \textbf{0.423} & 0.436 & 0.451 & 0.411 & 0.432 & 0.421 & 0.435 & 0.419 & 0.453& 0.436 & 0.503  \\
		
		\midrule
		
		\multirow{1}{*}{\textbf{ETTh2}} 
		& Avg  & \textbf{0.338} & \textbf{0.393} & 0.420 & 0.437 & 0.347 & 0.410 & 0.388 & 0.398  & 0.405 & 0.435 & 0.423 & 0.454 \\
		
		\midrule
		
		\multirow{1}{*}{\textbf{ETTm1}} 
		& Avg  & \textbf{0.409} & \textbf{0.431} & 0.465 & 0.510 & 0.417 & 0.441 & 0.436 & 0.473  & 0.450 & 0.487 & 0.468 & 0.515\\

		\midrule
		
		\multirow{1}{*}{\textbf{ETTm2}} 
		& Avg  & \textbf{0.247} & \textbf{0.301} & 0.301& 0.361 & 0.259 & 0.322 & 0.270 & 0.343   & 0.286 & 0.357 & 0.297 & 0.381\\

		\bottomrule
	\end{tabularx}
	\caption{IdealTSF of Ablation Studiese on ETT Datasets.}
	\label{t5}
\end{table*}

\section{Module-Specific Experiments}

\subsection{Feasibility Heatmap of Negative Sample Pre-training}

\begin{figure}[htbp]
	\centering
	{
		\includegraphics[width=1.6in]{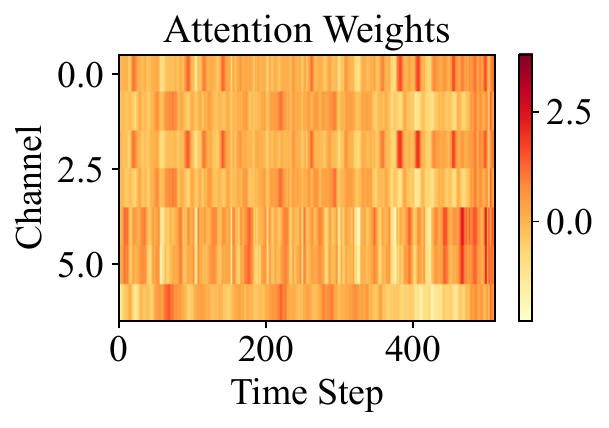}
	}
	{
		\includegraphics[width=1.6in]{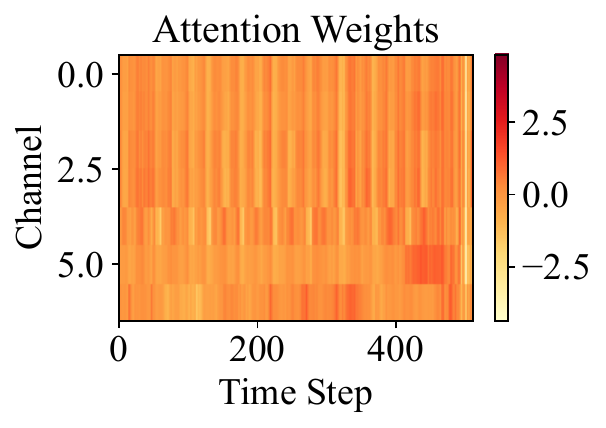}
	}\\
	
	\caption{(a) The left figure shows the attention heatmap of ETTh1 without negative sample pre-training. (b) The right figure shows the attention heatmap of ETTh1 after applying negative sample pre-training.}
	\label{f8}
\end{figure}

\begin{figure}[htbp]
	\centering
	{
		\includegraphics[width=1.5in]{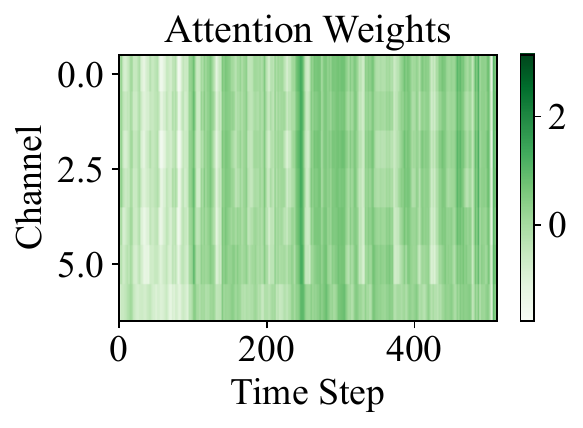}
	}
	{
		\includegraphics[width=1.6in]{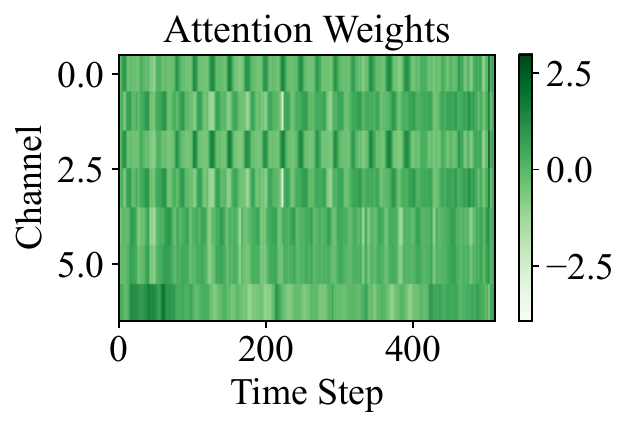}
	}\\
	
	\caption{(a) The left figure shows the attention heatmap of ETTh2 without negative sample pre-training. (b) The right figure shows the attention heatmap of ETTh2 after applying negative sample pre-training.}
	\label{f9}
\end{figure}

As shown in the attention heatmaps of the ETTh dataset in Figure~\ref{f8}–\ref{f9}, the left panel illustrates the attention distribution without negative sample pre-training, while the right panel displays the final attention weights after applying negative sample pre-training. It can be observed that, after pre-training, the attention distribution shifts from a uniform pattern to a focused concentration on specific time steps. This indicates that the model has learned meaningful temporal dependencies and is able to adaptively identify and focus on critical time points or intervals in the time series.

\subsection{ECOS Defense Performance Under Adversarial Attacks}
\begin{figure}[!t]
	\centering
	\includegraphics[width=1\linewidth]{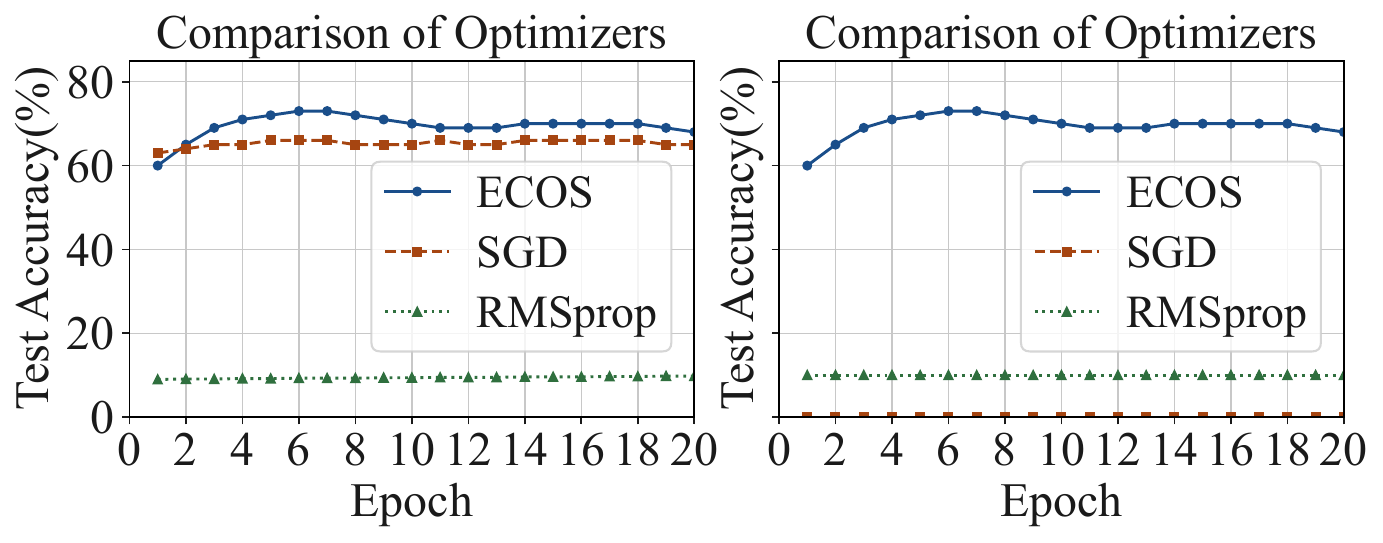}
	\caption{(a) Performance of IdealTSF under FGSM attack using different optimizers. (b) Performance of IdealTSF under PGD attack using different optimizers.}
	\label{f6}
\end{figure}

Just as ecosystems exhibit resilience to external disruptions, the robustness of an optimizer can be evaluated by its performance under adversarial inputs. The left plot of Figure~\ref{f6} illustrates that under FGSM attacks, the ECOS optimizer maintains higher accuracy and stability compared to SGD and RMSprop. Notably, RMSprop shows a significant performance drop when exposed to adversarial perturbations. The right plot of Figure~\ref{f6} further demonstrates that under the stronger PGD attack, only ECOS remains robust, while other optimizers suffer substantial performance degradation after the first few attack steps. These results highlight that ECOS is considerably more resilient to adversarial perturbations such as FGSM and PGD, making it particularly well-suited for scenarios involving risk, noise, or data anomalies.

\subsection{Effectiveness of the ECOS Optimizer}
\begin{figure}[!t]
	\centering
	\includegraphics[width=1\linewidth]{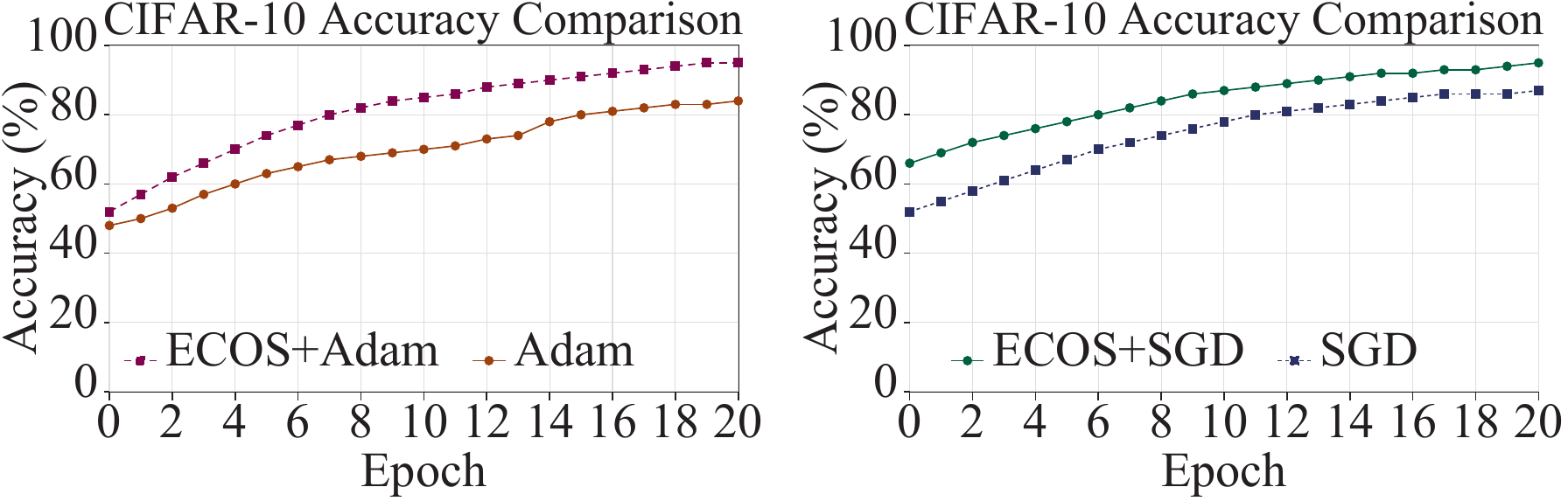}
	\caption{(a) The left figure shows the performance of the Adam and ECOS+Adam optimizers. (b) The right figure presents the performance of the SGD and ECOS+SGD optimizers.}
	\label{f7}
\end{figure}

To evaluate the effectiveness of the ECOS optimizer, we compare the performance of different optimization strategies—Adam vs. ECOS+Adam and SGD vs. ECOS+SGD—on the CIFAR-10 dataset, as shown in Figure~\ref{f7}. In the left panel, the accuracy of the standard Adam optimizer steadily increases with training epochs, ultimately reaching around 80\%. In contrast, ECOS+Adam exhibits a faster accuracy gain in the early stages and maintains higher stability in later epochs, achieving an accuracy of 90\%. In the right panel, the standard SGD optimizer shows relatively slow improvement, stabilizing at approximately 80\%, while ECOS+SGD significantly outperforms it, reaching nearly 90\% accuracy. Overall, the ECOS optimizer substantially enhances both training speed and accuracy on CIFAR-10, improving optimization performance when applied to either Adam or SGD.

\section{Conclusions}

This paper proposes a time series forecasting model, IdealTSF, which leverages imperfect data (negative samples) during the pre-training, training, and optimization phases to enhance model robustness. By incorporating negative sample pre-training, hybrid smoothing interpolation, and adversarial training, IdealTSF effectively handles missing and anomalous data, thereby improving adaptability to complex real-world scenarios. Under conditions of low data quality or high uncertainty, the model outperforms traditional approaches in both accuracy and generalization. Experimental results demonstrate the effectiveness of the proposed method. Future innovations and improvements will focus more on multi-dimensional and complex time-series datasets with strong practical significance, so as to further enhance the usability and generalization ability of IdealTSF in real-world business scenarios. Meanwhile, we will investigate robust training strategies under harsher data-quality conditions, including negative-sample construction that better matches real noise distributions.

\section{Appendix}
\bigskip 
\bigskip 
\section{Related Work}
\begin{table*}[t]
	\centering
	\caption{Long-Term Time Series Forecasting Performance of IdealTSF}
	
	\begin{tabularx}{\textwidth}{l c | *{2}{Y}  *{6}{Y}  *{8}{Y}}
		\toprule
		\textbf{} & \multicolumn{1}{c}{} & \multicolumn{2}{c|}{Ours} & \multicolumn{6}{c|}{2025}   & \multicolumn{8}{c}{2023-2024}  \\
		\midrule
		\textbf{Models} & \multicolumn{1}{c}{H} & \multicolumn{2}{c|}{IdealTSF} & \multicolumn{2}{c}{Twins*} & \multicolumn{2}{c}{-Mixer++}& \multicolumn{2}{c}{-KAN}& \multicolumn{2}{c}{SAM*} & \multicolumn{2}{c}{iTrans*} & \multicolumn{2}{c}{PatchTST} & \multicolumn{2}{c}{DLinear}  \\
		\cmidrule(lr){3-4} \cmidrule(lr){5-6} \cmidrule(lr){7-8} \cmidrule(lr){9-10} \cmidrule(lr){11-12} \cmidrule(lr){13-14} \cmidrule(lr){15-16} \cmidrule(lr){17-18}
		& & MSE & MAE & MSE & MAE & MSE & MAE & MSE & MAE & MSE & MAE & MSE & MAE & MSE & MAE & MSE & MAE \\
		\midrule
		\multirow{4}{*}{ETTh1}
		& 96 & \textbf{0.356} & \textbf{0.386} & {0.385} &{0.401} & 0.361 & 0.403 & 0.367 & 0.395 &0.381 &0.402 &0.386 &0.405 & 0.386 & 0.395 & 0.386 & 0.400 \\
		& 192 & \textbf{0.393} & \textbf{0.410} &{0.439} & {0.431} & 0.416 & 0.441 & 0.414 & 0.420 &0.409 &0.418 &0.441 &0.436  & 0.407 & 0.432 & 0.423 & 0.450  \\
		& 336 & \textbf{0.417} & \textbf{0.428} &0.480 &0.452  & 0.430 & 0.434 & 0.445 & 0.434 &0.423 &0.425 &0.487 &0.458  & 0.479 & 0.484 & 0.481 & 0.453  \\
		& 720 & \textbf{0.442} & \textbf{0.451} &{0.480} & {0.474} & {0.467} & \textbf{0.451} & 0.444 & 0.459 &0.427 &0.449 &0.503 &0.491  & 0.481 & 0.486 & 0.474 & 0.453  \\
		\midrule
		\multirow{4}{*}{ETTh2}
		& 96 & \textbf{0.270} & \textbf{0.342} &{0.292} &{0.345}  & 0.276 & 0.328 & 0.290 & 0.340  &0.295 &0.358 &0.297 &0.349 & 0.302 & 0.348 & 0.333 & 0.387  \\
		& 192 & \textbf{0.325} & \textbf{0.378} &{0.375} & {0.395} & 0.342 & 0.379 & 0.375 & 0.392  &0.340 &0.386 &0.380 &0.400 & 0.376 & 0.407 & 0.432 & 0.457  \\
		& 336 & \textbf{0.346} & \textbf{0.406} &{0.417} & {0.429} & 0.346 & 0.398 & 0.423 & 0.435  &0.350 &0.395 &0.428 &0.432& 0.420 & 0.439 & 0.594 & 0.581  \\
		& 720 & \textbf{0.410} & \textbf{0.445} & {0.406} & {0.430}  & 0.392 & 0.415 & 0.443 & 0.449  &0.391 &0.428 &0.427 &0.445 & 0.439 & 0.460 & 0.831 & 0.657  \\
		\midrule
		\multirow{4}{*}{ETTm1}
		& 96 & 0.325 & 0.379 &{0.325}& {0.364} & \textbf{0.310} & \textbf{0.334} & 0.322 & 0.361  &0.329 &0.363 &0.334 &0.368 & 0.336 & {0.371} & {0.320} & 0.372  \\
		& 192 & 0.385 & 0.409 & {0.372} &{0.390}  & \textbf{0.348} & \textbf{0.362} & 0.357 & {0.383}  &0.353 &0.378 &0.377 &0.391 & {0.367} & 0.391 & 0.414 & 0.410  \\
		& 336 & 0.431 & 0.458 & {0.406} & {0.412} & \textbf{0.376} & \textbf{0.391} & 0.382 & 0.401 &0.382 &0.394 &0.426 &0.420  & 0.420 & {0.410} & {0.413} & 0.413  \\
		& 720 & 0.496 & 0.479 & {0.467} & {0.448} & \textbf{0.440} & \textbf{0.423} & 0.445 & 0.435  &0.429 &0.418 &0.491 &0.459 & 0.439 & 0.474 & 0.453 & {0.453}  \\
		\midrule
		\multirow{4}{*}{ETTm2}
		& 96 & \textbf{0.157} & \textbf{0.238} &{0.173}& {0.256} &0.170& 0.245  & 0.174 & 0.255  &0.181 &0.274 &0.180 &0.264 & 0.175 & 0.218 & 0.193 & 0.255  \\
		& 192 & \textbf{0.202} & \textbf{0.275} & {0.239} & {0.300} & 0.229 & 0.291 & 0.239 & 0.299  &0.233 &0.306 &0.250 &0.309 & 0.241 & 0.282 & 0.284 & 0.362  \\
		& 336 & \textbf{0.275} & \textbf{0.324} & {0.298} & {0.339} & 0.303 & 0.343 & 0.301 & 0.340 &0.285 &0.338 &0.311 &0.348  & 0.305 & 0.364 & 0.369 & 0.427 \\
		& 720 & \textbf{0.356} & \textbf{0.370} &{0.397} & {0.397} & 0.373 & 0.399 & 0.395 & 0.396  &0.375 &0.390 &0.412 &0.407 & 0.400 & 0.400 & 0.554 & 0.522  \\
		\midrule
		\multirow{4}{*}{WTH}
		& 96 & \textbf{0.139} & \textbf{0.192} &{0.161} &{0.201}  & 0.155 & 0.205 & 0.162 & 0.208  &0.197 &0.249 &0.174 &0.214 & 0.177 & 0.218 & 0.196 & 0.255  \\
		& 192 & \textbf{0.176} & \textbf{0.230} &{0.211} &{0.248}  & 0.201 & 0.245 & 0.207 & 0.249  &0.235 &0.277 &0.221 &0.254 & 0.225 & 0.259 & 0.237 & 0.296  \\
		& 336 & \textbf{0.216} & \textbf{0.263} &{0.266} &{0.291}  & 0.237 & 0.265 & 0.263 & 0.290  &0.276  &0.304 &0.278 &0.296  & 0.293 & 0.297 & 0.283 & 0.359  \\
		& 720 & \textbf{0.271} & \textbf{0.309} &{0.347} & {0.343} & 0.312 & 0.334 & 0.338 & 0.340 &0.334 &0.342 &0.358 &0.347  & 0.348 & 0.345 & 0.381 & 0.487  \\
		\midrule
		\multirow{4}{*}{Traffic}
		& 96 & \textbf{0.348} & \textbf{0.244} &{0.382} &{0.260}  & 0.392 & 0.253 & - & -  &0.407 &0.292 &0.395 &0.268 & 0.544 & 0.395 & 0.650 & 0.396  \\
		& 192 & \textbf{0.355} & \textbf{0.243} &{0.392} & {0.267} & 0.402 & 0.258 & - & -  &0.415 &0.294 &0.417 &0.276 & 0.540 & 0.398 & 0.598 & 0.370  \\
		& 336 & \textbf{0.380} & \textbf{0.259} &{0.410} &{0.276}  & 0.428 & 0.263 & - & -  &0.421 &0.292 &0.433 &0.283 & 0.551 & 0.413 & 0.635 & 0.427  \\
		& 720 & \textbf{0.399} & \textbf{0.266} &{0.442} & {0.292} & 0.441 & 0.282 & - & -  &0.456 &0.311 &0.467 &0.302 & 0.586 & 0.375 & 0.645 & 0.394  \\
		\midrule
		\multirow{4}{*}{ECL}
		& 96 & \textbf{0.126} & \textbf{0.228} &0.139 &0.233  & 0.135 & 0.222 & 0.174 & 0.255  &0.155 &0.252 &0.148 &0.240 & 0.195 & 0.285 & 0.197 & 0.282  \\
		& 192 & \textbf{0.145} & \textbf{0.237} &0.158 &0.252  & 0.147 & 0.235 & 0.162 & 0.253  &0.168 &0.263 &0.162 &0.253 & 0.199 & 0.289 & 0.200 & 0.285  \\
		& 336 & \textbf{0.163} & \textbf{0.263} &0.172 &0.267  & 0.164 & 0.245 & 0.167 & 0.269  &0.183 &0.277 &0.167 &0.269 & 0.203 & 0.319 & 0.203 & 0.310  \\
		& 720 & \textbf{0.189} & \textbf{0.281} &0.200 &0.293  & 0.212 & 0.310 & 0.225 & 0.317  &0.219 &0.306 &0.225 &0.317 & 0.237 & 0.331 & 0.245 & 0.333  \\
		
		\bottomrule
	\end{tabularx}
	\footnotesize * indicates a former-based model , - indicates a Time-based model
	\label{At2}
\end{table*}

\subsection{Sudden disruptions in time series forecasting}
Sudden events—such as extreme occurrences, abnormal fluctuations, or abrupt regime shifts—pose significant challenges to model performance and forecasting accuracy in time series prediction. Traditional forecasting methods generally assume stationarity in trends and patterns, relying on the premise that future behavior can be inferred from historical regularities. However, such unexpected events often exhibit high nonlinearity and low predictability, which can dramatically increase forecasting errors. To address these challenges, numerous studies have incorporated anomaly detection mechanisms into time series forecasting frameworks. Statistical anomaly detection techniques, such as Z-score analysis and boxplot-based methods \cite{schmidl2022anomaly}, have been employed to identify outliers and improve predictive accuracy by removing or specially handling anomalous data points. In recent years, the rise of deep learning has introduced new perspectives on anomaly detection. Techniques such as Autoencoders and Generative Adversarial Networks (GANs) \cite{xu2022tgan} have demonstrated promising performance in detecting anomalies in complex temporal data. Another class of methods aims to directly model sudden events using stochastic frameworks such as jump process models or structural break models. For instance, Jump Diffusion Models \cite{chen2023imdiffusion} integrate classical Brownian motion with discrete jumps to capture abrupt fluctuations in prices or other time series signals. These models enable dynamic parameter adjustments in response to disruptive events, offering a principled approach to enhancing robustness under non-ideal conditions.

\subsection{Learning from Negative Samples}

Specific forms of undesirable behavioral data—commonly referred to as negative samples—have recently emerged as a focal point in time series analysis. Researchers have explored ways to leverage these negative samples to correct or suppress undesirable patterns through targeted modeling strategies \cite{he2019negative}. In the context of time series forecasting, He and Glass (2020) performed negative updates using signals derived from negative samples, aiming to suppress the generation of similar undesirable patterns during prediction. Similarly, \cite{lagutin2021implicit} introduced non-likelihood-based loss functions to penalize time series outputs with specific undesirable characteristics. Extending this line of work, \cite{li2019don} proposed enhancing the model’s ability to distinguish anomalous fluctuations by maximizing the predictive divergence between a student model and a pessimistic teacher. Although these approaches commonly treat negative samples as a source of corrective supervision, they often overlook the potentially informative signals embedded within them. In fact, negative samples may offer not only signals for suppression but also latent, beneficial information. This study aims to explore how negative samples can be incorporated into pre-training in the context of time series data, and how their synergistic integration with positive samples can enhance the extraction of critical features—ultimately improving both generalization and robustness.

\begin{table*}[t]
	\centering
	\caption{Dataset and model hyper-parameters along with training process configurations.}
	\begin{tabularx}{\textwidth}{l|*{9}{Y}|*{3}{Y}}
		\toprule
		& \multicolumn{9}{c}{\textbf{Model Hyper-parameter}} & \multicolumn{3}{c}{\textbf{Training Process}} \\ \cmidrule(r){2-10} \cmidrule(r){11-13}
		\textbf{Dataset} & $noise$ & $erase_l$ & $erase_p$ & $noise_{sc}$ &$l_{r}$ &$rho$ &$steps$ &$epsilon$ &$PGD$& \textbf{Loss} & \textbf{Batch Size} & \textbf{Epochs} \\ \midrule
		ETTh1  & 0.03 & 4-300 & 0.3 & 3  &1$e^{-3}$ &0.9 &3 &0.1 &0.5 & MSE & 32 & 30 \\ 
		ETTh2  & 0.03 & 4-300 & 0.3 & 3  &1$e^{-3}$ &0.9  &3 &0.1 &0.5 &MSE & 32 & 30 \\ 
		ETTm1  & 0.03 & 4-300 & 0.3  & 3  &1$e^{-3}$ &0.1  &3 &0.1 &0.5 &MSE & 32 & 30 \\ 
		ETTm2  & 0.03 & 4-300 & 0.3 & 3  &1$e^{-3}$ &0.1  &3 &0.1 &0.5 &MSE & 32 & 30 \\ 
		Weather  & 0.03 & 4-300 & 0.3  & 3  &1$e^{-3}$ &0.1  &3 &0.1 &0.5 &MSE & 32 & 30 \\ 
		Electricity  & 0.03 & 4-300 & 0.3  & 3  &1$e^{-3}$ &0.1  &3 &0.1 &0.5 &MSE & 32 & 30 \\ 
		Solar-Energy  & 0.03 & 4-300 & 0.3  & 3  &1$e^{-3}$  &0.1 &3 &0.1 &0.5 &MSE & 32 & 30 \\ 
		Traffic  & 0.03 & 4-300 & 0.3  & 3  &1$e^{-3}$ &0.1  &3 &0.1 &0.5 &MSE & 32 & 50 \\ 
		PEMS  & 0.03 & 4-300 & 0.3  & 3  &1$e^{-3}$ &0.1 &3 &0.1 &0.5 &MSE & 32 & 10 \\ 
		M4  & 0.03 & 4-300 & 0.3  & 3  &1$e^{-3}$ &0.1 &3 &0.1 &0.5 &SMAPE & 16 & 50 \\ \bottomrule
	\end{tabularx}
	\label{At1}
\end{table*}

\begin{table*}[ht]
	\centering
	\caption{Short-term forecasting performance on the PEMS dataset with prediction lengths of ${12, 24, 48, 96}$}
	\begin{tabularx}{\textwidth}{l c c c *{16}{Y}}
		\toprule
		
		\textbf{Models} & \multicolumn{1}{c}{(H)} & \multicolumn{2}{c}{IdealTSF}  & \multicolumn{2}{c}{Twins*}  & \multicolumn{2}{c}{itrans* } & \multicolumn{2}{c}{RLinear} & \multicolumn{2}{c}{PatchTST} & \multicolumn{2}{c}{Cross*} & \multicolumn{2}{c}{TiDE} & \multicolumn{2}{c}{TimesNet}  \\
		&  & MSE & MAE  & MSE & MAE & MSE & MAE & MSE & MAE & MSE & MAE & MSE & MAE & MSE & MAE  & MSE & MAE \\
		\midrule
		\multirow{4}{*}{\textbf{PEMS03}} 
		& 12 & \textbf{0.049} & \textbf{0.147} &0.065 &0.169 &0.069 &0.175  & 0.126 & 0.236 & 0.099 & 0.216 & 0.090 & 0.203 & 0.178 & 0.305 & 0.085 & 0.192 \\
		& 24 & \textbf{0.082} & \textbf{0.161} &0.086 &0.196 &0.097 &0.208  & 0.246 & 0.334 & 0.142 & 0.312 & 0.121 & 0.202 & 0.371 & 0.318 & 0.118 & 0.223  \\
		& 48 & \textbf{0.119} & \textbf{0.179} &0.121 &0.234 &0.131 &0.243  & 0.551 & 0.529 & 0.211 & 0.319 & 0.202 & 0.317 & 0.463 & 0.155 & 0.206 & 0.315  \\
		& 96 & \textbf{0.163} & \textbf{0.215} &0.165 &0.276 &0.168 &0.279  & 1.057 & 0.787 & 0.269 & 0.370 & 0.262 & 0.367 & 0.490 & 0.539 & 0.228 & 0.317  \\
		
		\midrule
		\multirow{4}{*}{\textbf{PEMS04}} 
		& 12 & \textbf{0.076} & \textbf{0.174} &0.077 &0.181 &0.081 &0.188 & 0.138 & 0.252 & 0.105 & 0.224 & 0.098 & 0.218 & 0.219 & 0.340 & 0.087 & 0.195 \\
		& 24 & \textbf{0.094} & \textbf{0.202} &0.095 &0.204 &0.099 &0.211 & 0.258 & 0.348 & 0.153 & 0.275 & 0.131 & 0.256 & 0.292 & 0.330 & 0.108 & 0.215 \\
		& 48 & \textbf{0.112} & \textbf{0.215}  &0.120 &0.231 &0.133 &0.247 & 0.572 & 0.544 & 0.229 & 0.339 & 0.205 & 0.336 & 0.409 & 0.478 & 0.136 & 0.250 \\ 
		& 96 & \textbf{0.142} & \textbf{0.258}  &0.150 &0.261 &0.172 &0.283 & 1.137 & 0.820 & 0.291 & 0.389 & 0.402 & 0.457 & 0.492 & 0.532 & 0.190 & 0.303 \\
		
		\midrule
		\multirow{4}{*}{\textbf{PEMS07}} 
		& 12 & \textbf{0.060} & {0.164}  &\textbf{0.060} &\textbf{0.158} &0.067 &0.167 & 0.118 & 0.235 & 0.095 & 0.207 & 0.094 & 0.200 & 0.173 & 0.304 & 0.082 & 0.181  \\
		& 24 & {0.096} & 0.209  &\textbf{0.079} &\textbf{0.181} &0.086 &0.189 & 0.242 & 0.341 & 0.150 & 0.262 & 0.139 & 0.247 & 0.271 & 0.383 & 0.101 & 0.204  \\
		& 48 & {0.132} & {0.233} &\textbf{0.104} &\textbf{0.209} &0.110 &0.214  & 0.562 & 0.541 & 0.253 & 0.340 & 0.311 & 0.369 & 0.446 & 0.495 & 0.134 & 0.238  \\
		& 96 & 0.182 & {0.275} &\textbf{0.132} &\textbf{0.236} &0.138 &0.244  & 1.096 & 0.795 & 0.346 & 0.404 & 0.396 & 0.442 & 0.628 & 0.577 & 0.181 & 0.279  \\
		
		\midrule		
		\multirow{4}{*}{\textbf{PEMS08}} 
		& 12 & {0.069} & {0.201}  &\textbf{0.075} &\textbf{0.174} &0.080 &0.183 & 0.133 & 0.247 & 0.168 & 0.232 & 0.165 & 0.214 & 0.227 & 0.343 & 0.112 & 0.212  \\
		& 24 & {0.108} & {0.212} &\textbf{0.106} &\textbf{0.206} &0.118 &0.221 & 0.249 & 0.343 & 0.224 & 0.281 & 0.215 & 0.260 & 0.318 & 0.409 & 0.141 & 0.238  \\
		& 48 & {0.175} & {0.258} &\textbf{0.167} &\textbf{0.258} &0.186 &0.265 & 0.569 & 0.544 & 0.321 & 0.354 & 0.315 & 0.355 & 0.497 & 0.510 & 0.198 & 0.283  \\
		& 96 & {0.229} & {0.281}  &\textbf{0.184} &\textbf{0.251} &0.221 &0.267 & 1.166 & 0.814 & 0.408 & 0.417 & 0.377 & 0.397 & 0.721 & 0.592 & 0.320 & 0.351  \\
		
		\bottomrule
	\end{tabularx}
	\footnotesize * indicates a former-based model
	\label{At3}
\end{table*}

\begin{table}[t]
	\centering
	\begin{tabular}{|l|c|c|}
		\hline
		\textbf{Model}      & \textbf{Parameters} & \textbf{MACs}   \\ \hline
		TimesNet            & 301.7M              & 1226.49G        \\ 
		Pyraformer          & 241.4M              & 0.80G           \\ 
		Informer            & 14.38M              & 3.93G           \\ 
		Autoformer          & 14.91M              & 4.41G           \\ 
		FiLM                & 14.91M              & 5.97G           \\ 
		FEDformer           & 20.68M              & 4.41G           \\ 
		PatchTST            & 1.5M                & 5.07G           \\ \hline
		IdealTSF (Ours)     & \textbf{1.2M}       & \textbf{198M} \\ \hline
	\end{tabular}
	\caption{Comparison of model parameters and MACs (Multiply-Accumulate Operations) across different models.}
	\label{At6}
\end{table}	

\begin{table*}[h!]
	\centering
	\caption{Model performance comparison across different metrics}
	\renewcommand{\arraystretch}{0.85} 
	\setlength{\tabcolsep}{3pt} 
	\begin{tabularx}{\textwidth}{ll*{17}{Y}}
		\toprule
		\textbf{Models} & & \textbf{IdealTSF} & \textbf{-Mixer} & \textbf{-Net} & \textbf{HiTS} & \textbf{BEATS} & \textbf{SCINet} & \textbf{Patch*} & \textbf{MICN} & \textbf{FiLM} & \textbf{LightTS} & \textbf{DLinear}   \\ 
		& & (Ours) & (2025) & (2023a) & (2023) & (2019) & (2022a) & (2023) & (2023) & (2022a) & (2022) & (2023)    \\ \midrule
		\multirow{3}{*}{\textbf{Yearly}} 
		& SMAPE & \textbf{13.225} &13.358 & 13.387 & 13.418 & 13.436 & 18.605 & 16.463 & 25.022 & 17.431 & 14.247 & 16.965    \\
		& MASE & \textbf{0.808} &3.021 & 2.996 & 3.045 & 3.043 & 4.471 & 3.967 & 7.162 & 4.043 & 3.109 & 4.283   \\
		& OWA & \textbf{0.742} &0.789 & 0.786 & 0.793 & 0.794 & 1.132 & 1.003 & 1.667 & 1.042 & 0.827 & 1.058   \\ \midrule
		\multirow{3}{*}{\textbf{Quarterly}} 
		& SMAPE & {13.849} &\textbf{10.058} & 10.100 & 10.202 & 10.124 & 14.871 & 10.644 & 15.214 & 12.925 & 11.364 & 12.145    \\
		& MASE & \textbf{0.773} &1.170 & 1.182 & 1.194 & 1.169 & 2.054 & 1.278 & 1.963 & 1.664 & 1.328 & 1.520   \\
		& OWA & \textbf{0.702} &0.883 & 0.890 & 0.899 & 0.886 & 1.203 & 1.140 & 1.445 & 1.407 & 1.193 & 1.106   \\ \midrule
		\multirow{3}{*}{\textbf{Monthly}} 
		& SMAPE & {13.269} &12.717 & \textbf{12.670} & 12.791 & 12.677 & 14.925 & 13.399 & 16.943 & 15.407 & 12.514 & 14.260   \\
		& MASE & \textbf{0.709} &0.933 & 0.933 & 1.131 & 1.031 & 1.445 & 1.256 & 1.944 & 1.303 & 1.145 & 1.397   \\
		& OWA & \textbf{0.603} &0.879 & 0.878 & 0.899 & 0.880 & 1.027 & 0.949 & 1.265 & 1.144 & 0.952 & 1.019  \\ \midrule
		\multirow{3}{*}{\textbf{Others}} 
		& SMAPE & {5.125} &\textbf{4.845} & 4.891 & 5.061 & 4.925 & 16.655 & 6.558 & 41.985 & 7.134 & 5.880 & 6.709   \\
		& MASE & \textbf{1.996} &3.217 & 3.302 & 3.216 & 3.393 & 4.511 & 4.154 & 62.734 & 4.364 & 4.064 & 4.953    \\
		& OWA & \textbf{0.812} &1.017 & 1.035 & 1.043 & 1.053 & 1.263 & 1.401 & 3.141 & 1.353 & 1.340 & 1.187   \\ 
		\bottomrule
	\end{tabularx}
	\footnotesize * indicates TST , - indicates Times
	\label{At4}
\end{table*}

\subsection{Optimization}
Many time series forecasting models exhibit strong performance on training datasets but tend to overfit when deployed on future test data. This overfitting often stems from excessive reliance on historical patterns, leading to poor generalization across unseen time steps. As illustrated in Figure 1, time series data frequently contain noise or instability, particularly in domains such as finance and sensor analytics. In such environments, model gradients may become highly sharp, making the training process overly sensitive to fluctuations and reducing predictive performance on unseen data.Widely adopted optimizers—such as Stochastic Gradient Descent (SGD) with momentum \cite{nesterov2013gradient}, Adam \cite{kingma2014adam}, RMSProp \cite{zou2019sufficient}, and other variants \cite{duchi2011adaptive, dozat2016incorporating, martens2015optimizing}—typically rely on the direction of local gradients for parameter updates. However, in complex, non-convex loss landscapes, especially those with sharp curvature, these methods are prone to becoming trapped in narrow local minima. As a result, they often fail to converge to flatter optima that are known to generalize better across varying data distributions.

In contrast to previous approaches, the proposed IdealTSF framework enhances robustness by injecting perturbations during training to generate adversarial samples, which improve the model’s adaptability to sudden anomalies and diverse imperfections. Moreover, IdealTSF integrates fine-tuning techniques inspired by large language models (LLMs), enabling faster convergence toward flat and generalizable solutions.

\begin{table*}[t]
	\centering
	\caption{IdealTSF of Ablation Studiese on ETT Datasets}
	\begin{tabularx}{\linewidth}{c|c|*{12}{>{\centering\arraybackslash}X}}
		\toprule
		\multicolumn{2}{c|}{\textbf{Models}}  & \multicolumn{2}{c|}{\textbf{IdealTSF}} & \multicolumn{2}{c|}{\textbf{w/o neg}} & \multicolumn{2}{c|}{\textbf{w/o Pos}} & \multicolumn{2}{c|}{\textbf{w/o ECOS}}   & \multicolumn{2}{c|}{\textbf{w/o Pos+ECOS}} & \multicolumn{2}{c|}{\textbf{w/o neg+ECOS}} \\
		\multicolumn{2}{c|}{\textbf{Metric}} & \textbf{MSE} & \textbf{MAE} & \textbf{MSE} & \textbf{MAE} & \textbf{MSE} & \textbf{MAE} & \textbf{MSE} & \textbf{MAE} & \textbf{MSE} & \textbf{MAE} & \textbf{MSE} & \textbf{MAE} \\
		\midrule
		\multirow{4}{*}{\textbf{ETTh1}} 
		& 96  & \textbf{0.356} & \textbf{0.386} & 0.382 & 0.410 & 0.362 & 0.399 & 0.371 & 0.401 & 0.388 & 0.431& 0.399 & 0.491  \\
		& 192 & \textbf{0.393} & \textbf{0.410} & 0.419 & 0.422 & 0.401 & 0.418 & 0.401 & 0.401 & 0.429 & 0.442& 0.439 & 0.499 \\
		& 336 & \textbf{0.417} & \textbf{0.428} & 0.452 & 0.475 & 0.429 & 0.434 & 0.431 & 0.459 & 0.449 & 0.492& 0.466 & 0.521\\
		& 720 & \textbf{0.442} & \textbf{0.466} & 0.492 & 0.498 & 0.450 & 0.477 & 0.481 & 0.479 & 0.409 & 0.442& 0.439 & 0.499\\
		
		\midrule
		
		\multirow{4}{*}{\textbf{ETTh2}} 
		& 96  & \textbf{0.270} & \textbf{0.342} & 0.322 & 0.395 & 0.279 & 0.353 & 0.299 & 0.360  & 0.303 & 0.401 & 0.321 & 0.419 \\
		& 192 & \textbf{0.325} & \textbf{0.378} & 0.418 & 0.409 & 0.329 & 0.399 & 0.381 & 0.394  & 0.400 & 0.407 & 0.411 & 0.421 \\
		& 336 & \textbf{0.346} & \textbf{0.406} & 0.439 & 0.443 & 0.361 & 0.421 & 0.411 & 0.406 & 0.450 & 0.477 & 0.481 & 0.479 \\
		& 720 & \textbf{0.410} & \textbf{0.445} & 0.499 & 0.502 & 0.417 & 0.466 & 0.459 & 0.433  & 0.468 & 0.455 & 0.479 & 0.497 \\
		
		\midrule
		
		\multirow{4}{*}{\textbf{ETTm1}} 
		& 96  & \textbf{0.325} & \textbf{0.379} & 0.399 & 0.451 & 0.329 & 0.387 & 0.373 & 0.400   & 0.383 & 0.407 & 0.400 & 0.459\\
		& 192 & \textbf{0.385} & \textbf{0.409} & 0.430 & 0.489 & 0.396 & 0.423 & 0.401 & 0.455   & 0.420 & 0.470 & 0.451 & 0.489\\
		& 336 & \textbf{0.431} & \textbf{0.458} & 0.498 & 0.521 & 0.442 & 0.467 & 0.471 & 0.499   & 0.487 & 0.519 & 0.499 & 0.530\\
		& 720 & \textbf{0.496} & \textbf{0.479} & 0.533 & 0.578 & 0.502 & 0.488 & 0.499 & 0.539   & 0.509 & 0.551 & 0.522 & 0.581\\
		
		\midrule
		
		\multirow{4}{*}{\textbf{ETTm2}} 
		& 96  & \textbf{0.157} & \textbf{0.238} & 0.199& 0.307 & 0.169 & 0.266 & 0.183 & 0.299   & 0.194 & 0.302 & 0.199 & 0.309\\
		& 192 & \textbf{0.202} & \textbf{0.275} & 0.268 & 0.355 & 0.216 & 0.320 & 0.249 & 0.344   & 0.269 & 0.362 & 0.287 & 0.387\\
		& 336 & \textbf{0.275} & \textbf{0.324} & 0.335 & 0.376 & 0.281 & 0.325 & 0.291 & 0.346  & 0.321 & 0.369 & 0.333 & 0.399\\
		& 720 & \textbf{0.356} & \textbf{0.370} & 0.404 & 0.406 & 0.370 & 0.377 & 0.358 & 0.386  & 0.360 & 0.398 & 0.371 & 0.430\\
		
		\bottomrule
	\end{tabularx}
	\label{At5}
\end{table*}

\section{Experiments}

\textbf{Related to the experiment} In the main text, due to space limitations, we used the average values to present both the long - term time series prediction and the short - term time series prediction. For the short - term time series prediction, in order to prove the resistance and robustness of our model to real - world noise, we conducted experiments on all baseline models by randomly adding noise ranging from 5\% to 35\%. Meanwhile, to ensure fairness, we presented the complete experimental data without adding noise in this part of the appendix.

\textbf{Metric details} For long-term forecasting, we adopt mean squared error (MSE) and mean absolute error (MAE) as evaluation metrics. In the case of short-term forecasting, we follow the evaluation protocol of SCINet , which includes the MAE coefficient, mean absolute percentage error (MAPE), and root mean squared error (RMSE). For the M4 dataset, we follow the N-BEATS methodology  and employ symmetric mean absolute percentage error (SMAPE), mean absolute scaled error (MASE), and overall weighted average (OWA) as evaluation metrics. It is worth noting that OWA is a specific metric used in the M4 competition.

\textbf{Environment} All experiments in this study were implemented using PyTorch (Paszke et al., 2019) and conducted on a single NVIDIA RTX 4060 8GB GPU and NVIDIA A800 80GB GPU.

\subsection{Architecture and Training Parameters}
To demonstrate the feasibility and reproducibility of the proposed model, Table~\ref{At1} provides detailed information on the model's hyperparameters and training configurations across different datasets and settings. Specifically: $\alpha$ denotes the magnitude of perturbation; $steps$ refers to the number of steps in multi-step updates; $\epsilon$ is the maximum allowed perturbation used in adversarial sample generation; $PGD$ indicates the step size for PGD-based adversarial training; $incremental$ specifies whether incremental training is enabled; $low_{order}$ determines whether low-order updates (i.e., simplified gradient computation) are used; $noise$ defines the standard deviation of default Gaussian noise; $erase_{l}$ is the maximum length of random deletion; $erase_{p}$ represents the probability of random deletion; $noise_{sc}$ lists the intensity of noise at different scales; and $l_{r}$ denotes the learning rate used by the optimizer.

\subsection{Long-Term Time Series Forecasting}
Table~\ref{At2} presents a comparison between IdealTS and other baseline models across four datasets. The results show that IdealTS consistently ranks among the top two performers on all datasets, often achieving or approaching state-of-the-art performance. Bold values indicate the best performance among all baselines. The datasets cover a wide range of time series with varying frequencies, numbers of variables, and real-world scenarios. In particular, IdealTS significantly outperforms models such as TimeKAN and PatchTST, reducing the MSE by approximately 10\% on the Weather dataset and by around 8\% on the ETT dataset. Notably, even on datasets with lower predictability such as Traffic and ECL, IdealTS still delivers strong performance, further demonstrating its robustness and generalizability.

\subsection{Short-Term Time Series Forecasting}
IdealTS also performs remarkably well in both multivariate and univariate short-term forecasting tasks (Tables~\ref{At3} \ref{At4}). On the multivariate PEMS benchmark, despite the presence of complex spatiotemporal dependencies, IdealTS outperforms advanced models such as PatchTST and DLinear, demonstrating strong adaptability. In the M4 univariate dataset, which includes a wide range of sampling frequencies, IdealTS consistently achieves the best performance across all frequencies, confirming its capability to model complex temporal dynamics and its strong generalization ability.

\subsection{Efficiency Advantages of IdealTSF}
Table~\ref{At6} reports the number of trainable parameters and multiply-accumulate operations (MACs) for various TSF models on the electricity dataset, using a look-back window of 96 and a forecasting horizon of 720. The table clearly demonstrates the superior efficiency of IdealTSF compared to other baseline models. Among deep learning models employing attention-based architectures, IdealTSF requires significantly fewer parameters. Despite its compact size, IdealTSF consistently achieves competitive results, making it an attractive choice for time series forecasting tasks. IdealTSF demonstrates that it is possible to achieve state-of-the-art or near state-of-the-art performance while significantly reducing parameter overhead, making it a compelling solution for deployment in resource-constrained environments.

\subsection{Ablation Study}
To evaluate the effectiveness of each component in IdealTSF, we conducted ablation studies by systematically removing individual modules (denoted as w/o). Table~\ref{At5} presents detailed results and corresponding analysis. Specifically, "w/o neg" refers to the removal of the negative sample pre-training module, "w/o pos" denotes the removal of the positive sample generation process, and "w/o ECOS" indicates the exclusion of the Ecosystem Optimizer. "w/o ECOS" indicates the model variant without the Ecosystem Optimizer. "w/o Pos+ECOS" removes both the positive sample training and ECOS, while "w/o Neg+ECOS" removes both negative sample pre-training and ECOS. From the MSE and MAE metrics across multiple datasets (ETTh1, ETTh2, ETTm1, and ETTm2) and forecast horizons (Avg), it is evident that IdealTSF consistently outperforms the ablated versions, demonstrating superior forecasting accuracy and stability. These results confirm the contribution of each module to the overall performance and robustness of the model.

\section{Mathematical Derivation and Feasibility Proof of the Ecosystem Optimizer (ECOS)}

\subsection{Optimization Objective and Framework of ECOS}

Suppose we have a target function (loss function) $ L(\theta) $, where $ \theta $ represents the model parameters. The objective is to minimize the loss:
$
\theta^* = \arg\min_\theta L(\theta)
$

However, traditional optimization methods (such as SGD or Adam) focus solely on minimizing the loss function, while ignoring the model's sensitivity to external factors. The goal of ECOS is to enhance the model's robustness to perturbations by considering sharp regions in the loss landscape and potential external disturbances. 

ECOS searches for flat regions through the following two-step update process:

\begin{itemize} \item \textbf{Phase 1 (Flat Region Exploration)}: At the current parameter point $ \theta_t $, compute a perturbation $ \Delta \theta_t $ that guides the model in the direction of a "local maximum" in the loss function. \item \textbf{Phase 2 (Recovery and Base Optimization)}: Restore the parameters to their original state, and then update them using a base optimizer such as SGD or Adam. \end{itemize}

In the following sections, we will provide a step-by-step mathematical derivation of these two phases.

\subsection{Phase One: Perturbation Update}

First, we define what constitutes a sharp region, namely the model's sensitivity to parameter perturbations. A sharp region in the model can be characterized by the maximum variation of the loss function within a local neighborhood in the parameter space:
$
\text{sharpness}(\theta) = \sup_{\|\Delta \theta\|_2 \leq \rho} \left| L(\theta + \Delta \theta) - L(\theta) \right|
$
Here, $ \rho $ controls the magnitude of the perturbation, and $ \Delta \theta $ denotes the direction of the perturbation.

The objective of the first phase of ECOS is to find a perturbation $ \Delta \theta $ such that the perturbed parameters $ \theta + \Delta \theta $ lead to a maximized increase in the loss function, thereby enhancing the model's robustness to external disturbances. In other words, ECOS ascends toward the local maxima of the loss landscape to avoid regions that are overly sensitive to perturbations during optimization.

\subsubsection{Computing the Perturbation Direction}

First, ECOS computes the perturbation direction based on gradient information. For a given parameter $ \theta $, the gradient of the loss function is denoted as $ \nabla_\theta L(\theta) $. The perturbation $ \Delta \theta $ is chosen to align with the gradient direction, with its magnitude controlled by the perturbation size $ \rho $. Specifically, ECOS computes a perturbation as follows:

$
\Delta \theta_t = \rho \cdot \frac{\nabla_\theta L(\theta_t)}{\|\nabla_\theta L(\theta_t)\|_2}
$

This corresponds to the unit vector in the direction of the gradient, scaled by a factor $ \rho $ (which controls the perturbation magnitude). This step ensures that the perturbation remains moderate, preventing unreasonable deviations in the parameter values.

\subsubsection{Why Should the Perturbation Magnitude Be Controlled?}

To maintain optimization stability, it is crucial to prevent the perturbation from being excessively large; otherwise, it may cause the parameters to shift significantly, leading to instability in the optimization process. In practice, ECOS constrains the perturbation magnitude to prevent the model from becoming overly sensitive during training. Specifically, the perturbation size $ \rho $ is determined by balancing the gradient magnitude and the scale of perturbation, thereby ensuring robust optimization. This strategy helps the model avoid overreacting to local noise or minor fluctuations.

\subsection{Stage Two: Parameter Restoration and Base Optimization}
After applying the perturbation in the first stage, the second stage of ECOS involves restoring the parameters to their original state and performing base optimization (e.g., SGD or Adam). The objective of this step is to revert the perturbed parameters and then conduct a standard gradient update step. This ensures stable optimization that continues in the direction of minimizing the loss function. Moreover, this process typically involves several small-step updates and forward passes.

\subsubsection{Parameter Restoration}
After the perturbation update in the first stage, the parameters become:
$
\theta_t' = \theta_t + \Delta \theta_t
$

In the second stage, the parameters are restored to their original position $ \theta_t $:
$
\theta_t = \theta_t' - \Delta \theta_t
$
The purpose of this restoration step is to eliminate the perturbation introduced in the first stage, thereby ensuring that excessive bias does not accumulate during the optimization process.

\subsubsection{Base Optimization Step}
After restoring the parameters, we apply a standard optimization update using a base optimizer (such as Adam or SGD). The update is given by:
$
\theta_{t+1} = \theta_t - \eta \nabla_\theta L(\theta_t)
$

At this stage, the model is no longer affected by perturbations, so the optimization step proceeds as a standard update based on the current gradient and learning rate.

\subsection{Incorporation of Adversarial Training}

In addition to perturbation-based updates, ECOS incorporates adversarial training to further enhance model robustness by generating adversarial examples. These adversarial examples are created by applying small perturbations to the input sample $ x $, with the goal of encouraging the model to remain robust against such perturbations.

\subsubsection{Objective of Adversarial Training}

The objective of adversarial training is to minimize the model's sensitivity to perturbations. Given an input sample $ x $ and its corresponding label $ y $, an adversarial sample $ x_{adv} $ is generated for loss computation. Common techniques for generating $ x_{adv} $ include the Fast Gradient Sign Method (FGSM) and Projected Gradient Descent (PGD).

For a given loss function $ \mathcal{L} $, the adversarial sample $ x_{adv} $ is updated using the following rule:
$
x_{adv} = x + \alpha \cdot \text{sign}(\nabla_x \mathcal{L}(f_\theta(x), y))
$
Here, $ \alpha $ denotes the step size, and $ \text{sign}(\nabla_x \mathcal{L}) $ represents the sign of the gradient of the loss function with respect to the input $ x $.

\subsubsection{Mathematical Interpretation of Adversarial Training}

Through adversarial training, the model is not only optimized to minimize loss on the training set, but also explicitly encouraged to maintain robustness under adversarial perturbations. Assuming the training is performed on adversarial samples $ x_{adv} $, the objective becomes:
$
\mathbb{E}_{(x, y) \sim p(x, y)} [ \mathcal{L}(f_\theta(x_{adv}), y) ]
$

The key to adversarial training lies in generating adversarial examples that minimize the performance gap between clean and perturbed inputs, thereby enhancing the model’s robustness.

\subsection{Summary and Feasibility Justification}

Having derived each component of ECOS in detail, we now summarize its overall feasibility as follows:

\begin{itemize} 
	\item \textbf{Flat Region Exploration}: In the first stage, ECOS computes a perturbation $ \Delta \theta $ and updates along the gradient direction, effectively “climbing” toward the local maxima of the loss surface. This increases robustness against perturbations. The perturbation magnitude $ \rho $ is carefully controlled to prevent instability caused by excessive updates.
	\item \textbf{Multi-Step Fine-Tuning}: After restoring the perturbation, ECOS performs standard gradient updates using a base optimizer (e.g., Adam or SGD), along with multiple-step probing to ensure the optimization trajectory remains stable and convergent.
	\item \textbf{Adversarial Training}: By generating and training on adversarial examples, ECOS further strengthens the model’s robustness to input perturbations and mitigates overfitting to the training data, thereby improving generalization.
\end{itemize}

\section{Discussion}
In this study, we propose the IdealTSF model for time series forecasting, targeting the common challenges posed by imperfect data such as missing values and anomalies in real-world scenarios. While our experiments demonstrate that incorporating negative samples—i.e., imperfect or anomalous data points—can significantly enhance forecasting robustness, an important direction for future work is to develop more principled strategies to balance positive and negative samples during training. In practical applications such as energy consumption forecasting, financial risk monitoring, or industrial process control, the frequency and severity of anomalies can vary greatly across domains and over time. Future research could therefore explore adaptive reweighting or curriculum-style sampling, where the contribution of negative samples is dynamically adjusted based on the model’s current robustness and the observed anomaly patterns in the deployment environment.

A key limitation of the current IdealTSF design is its potential to overfit the specific form of negative samples used during training. Since these negative samples are often generated by predefined perturbation rules or adversarial attacks (e.g., FGSM, PGD), the model may become overly specialized to these particular perturbation patterns, which may not fully match the complex, heterogeneous anomalies encountered in real-world systems. Future work could investigate more diverse and realistic negative-sample generation mechanisms, such as learning a generative model of anomalies from historical incident logs, or leveraging simulation-based perturbations that mimic domain-specific failure modes. Additionally, regularization techniques tailored to negative-sample learning—such as consistency constraints across different perturbation strengths or multi-view augmentations—may further reduce the risk of overfitting to any single type of noise.

In many application domains, anomalies and data imperfections are not arbitrary but follow domain-specific patterns (e.g., sensor drift in manufacturing, reporting delays in healthcare, or holiday effects in retail sales). An exciting future direction is to incorporate domain knowledge into the construction and utilization of negative samples in IdealTSF. For instance, domain experts can help define realistic perturbation templates, constraints on physically plausible trajectories, or rules governing how certain external events affect the time series. By embedding such knowledge into the negative-sample generation and training pipeline, IdealTSF can better distinguish between harmful anomalies and benign irregularities such as regime shifts or structural breaks, thereby improving both accuracy and interpretability in real deployments.

Real-world time series forecasting systems often operate in non-stationary environments, where data distributions evolve over time due to concept drift, system upgrades, or changing user behaviors. Extending IdealTSF to an online or continual learning framework represents another promising direction. In such a setting, the model could continuously update its representation of both positive and negative samples as new data arrives, and selectively forget outdated negative-sample patterns that no longer reflect the current system behavior. Furthermore, in privacy-sensitive domains such as finance and healthcare, federated variants of IdealTSF could be developed to learn from imperfect time series across multiple organizations without centralizing raw data, while still benefiting from shared patterns of anomalies and negative samples. Moreover, integrating the uncertainty estimates and robustness indicators produced by IdealTSF into downstream decision rules can help practitioners design more reliable alarm thresholds, risk scores, and intervention strategies that are explicitly aware of data imperfections.

Finally, although IdealTSF already shows substantial improvements in handling imperfect time series by jointly leveraging positive and negative samples, future research can further extend this paradigm to multivariate and multimodal forecasting scenarios. In many real-world systems, temporal signals are accompanied by auxiliary information such as graphs (e.g., sensor networks, transportation networks), textual logs, or images. Designing negative samples that simultaneously perturb multiple modalities—while maintaining cross-modal consistency—could reveal richer structural weaknesses of the model and lead to more robust representations. Exploring such extensions may open up new avenues for applying IdealTSF-like frameworks to domains such as smart grids, intelligent transportation, and complex cyber–physical systems.

\section*{Acknowledgements}
This work was supported in part by the following: the Joint Fund of the National Natural Science Foundation of China under Grant Nos. U24A20328, U24A20219, the Youth Innovation Technology Project of Higher School in Shandong Province under Grant No. 2023KJ212, the National Natural Science Foundation of China under Grant No. 62272281, the Special Funds for Taishan Scholars Project under Grant No. tsqn202306274, and the Natural Science Foundation of Shandong Province under Grant No. ZR20250C712.

\bibliography{aaai2026}

\end{document}